\documentclass[3p,twocolumn]{elsarticle}
\usepackage{tabularx,booktabs}
\usepackage{booktabs}
\usepackage{longtable}
\usepackage{multirow}
\usepackage{multicol}
\usepackage{graphicx}
\usepackage{amsmath}
\usepackage{verbatim}
\usepackage{amssymb}
\usepackage{amsmath}
\usepackage{subfig}
\usepackage{threeparttable}
\usepackage{diagbox}
\usepackage{hyperref,balance}

\begin{document}

\begin{frontmatter}

\title{Efficient Screening of Diseased Eyes based on Fundus Autofluorescence Images using Support Vector Machine}

\author[mymainaddress]{Shanmukh Reddy Manne\corref{mycorrespondingauthor}}
\ead{ee14resch11006@iith.ac.in}

\author[upits]{ Kiran Kumar Vupparaboina}

\author[mymainaddress]{ Gowtham Chowdary Gudapati}

\author[mymainaddress]{ Ram Anudeep Peddoju}

\author[mymainaddress]{ Chandra Prakash Konkimalla}

\author[lvpei]{Abhilash Goud}

\author[lvpei]{Sarforaz Bin Bashar}

\author[upits]{Jay Chhablani}

\author[mymainaddress]{ Soumya Jana}
\cortext[mycorrespondingauthor]{Corresponding author}


\address[mymainaddress]{Indian Institute of Technology Hyderabad, India}
\address[upits]{University of Pittsburgh, USA}
\address[lvpei]{L V Prasad Eye Institute, Hyderabad, India}

\begin{abstract}
\textbf{Background and Objective:} A variety of vision ailments are associated with geographic atrophy (GA) in the foveal region of the eye. In current clinical practice, the ophthalmologist manually detects potential presence of such GA based on fundus autofluorescence (FAF) images, and hence diagnoses the disease, when relevant. 
However, in view of the general scarcity of ophthalmologists relative to the large number of subjects seeking eyecare, especially in remote regions, it becomes imperative to develop methods to direct expert time and effort to medically significant cases. Further, subjects from either disadvantaged background or remote localities, who face considerable economic/physical barrier in consulting trained ophthalmologists, tend to seek medical attention only after being reasonably certain that an adverse condition exists. To serve the interest of both the ophthalmologist and the potential patient, we plan a screening step, where healthy and diseased eyes are algorithmically differentiated with limited input from only optometrists who are relatively more abundant in number.
\textbf{Methods:} We proposed a semi-automated framework where an early treatment diabetic retinopathy study (ETDRS) grid is  placed by an optometrist on each FAF image, based on which sectoral statistics are automatically collected. Using such statistics as features, healthy and diseased eyes are proposed to be classified by training an algorithm using available medical records. In this connection, we demonstrate the efficacy of support vector machines (SVM). Specifically, we consider SVM with linear as well as radial basis function (RBF) kernel, and observe satisfactory performance of both variants. 
\textbf{Results:} We considered FAF images acquired from 61 healthy eyes and 79 diseased eyes, in developing the proposed algorithm. As a result, we observe that SVM with RBF kernel has slight superiority over its linear version, in  terms  of  classification  accuracy (90.55\% against 89.88\% at a standard training-to-test ratio of 80:20), and practical class-conditional costs.
\textbf{Conclusion:} 
The proposed SVM-based screening system is efficient in aiding ophthalmologists and could be attractive from another practical consideration. Preliminary investigations suggest that an efficient disease progress monitoring system could potentially be developed also using an SVM platform, allowing cost savings due to shared technology.
\end{abstract}




\end{frontmatter}


\section{INTRODUCTION}
  \begin{figure*}[!t]
\centering
    \includegraphics[scale=0.22]{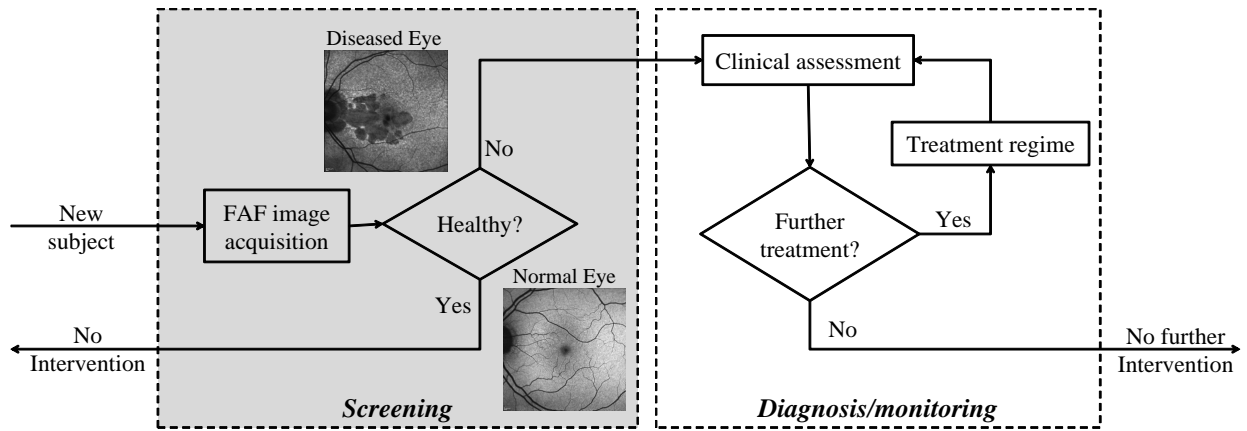}
\caption{An end-to-end framework for screening and diagnosis/treatment monitoring system. }
\label{fig:Intro}
\end{figure*}
Medical imaging plays a crucial role in managing ophthalmic diseases. Among current modalities \cite{yolcu2014imaging}, fundus autofluorescence (FAF) imaging presents a relatively new technology that exploits the innate fluorescence of the retinal pigment epithelium (RPE), and has proven attractive by not requiring injection of artificial dye \cite{nickla2010multifunctional}. Specifically, the retina is illuminated with blue light which causes the naturally occurring pigment lipofuscin to fluoresce \cite{schmitz2008fundus}. Consequently, the spatial distribution of intensity in the captured grey-scale FAF image indicates that of lipofuscin. Inspecting such distribution, the ophthalmologist is able to visualize the status of the retina (especially, retinal pigment epithelium), and hence diagnose/monitor various ophthalmic diseases. More precisely, for disease diagnosis, the physician generally hypothesizes a disease while inspecting a FAF image, and subsequently confirms it on the basis of quantitative indicators. For follow-up subjects, such indicators are used to monitor disease course. In either scenario, the associated clinical assessment provides the basis for deciding further treatment, if required. \\

To facilitate such clinical assessment, significant effort has been directed at  developing analysis tools specific to diseases, such as age-related macular degeneration (AMD) and Stargardt disease (STGD), an inherited ailment characterised by macular degeneration \cite{AMD_STGD}. Such tools are based on disease-specific quantitative indicators, whose discovery has also remained an important focus. Indeed, indicators based on sectoral statistics collected from multi-segment grids centered at the fovea have been reported for various diseases including  STGD \cite{Sect_Stargardt}, AMD \cite{Retina2013}  \cite{schmitz2008evaluation}, central serous chorioretinopathy (CSCR) \cite{zola2018evolution}, and retinitis pigmentosa (RP) \cite{jolly2016novel}. Further, owing to its high prevalence, special attention has been devoted to macular degeneration, studying variations in FAF phenotypes with disease progression \cite{schmitz2009fundus}, algorithmically quantifying associated geographic atrophy (GA)  \cite{schmitz2011semiautomated}, characterizing such GA \cite{fleckenstein2011fundus} \cite{holz2007progression}, and investigating factors influencing progression of such GA \cite{fleckenstein2018progression}. 
Quantification related to RP has also received significant attention \cite{schuerch2017quantifying}. 

Yet, advanced clinical assessment tools alone may not always ensure improved clinical outcome. To appreciate this, consider subjects from economically disadvantaged background and/or remote regions, who face considerable economic/physical barrier in consulting trained ophthalmologists. In those circumstances, subjects tend to seek medical attention only after being reasonably certain that an adverse condition exists. Specifically, as shown in Figure \ref{fig:Intro}, such subjects seek a locally accessible economical screening service. If diseased condition is indicated in the process, they become more likely to undertake the necessary yet arduous steps to obtain diagnosis, treatment and treatment response monitoring, as needed. In this context, accurate screening assumes a crucial role in ensuring desired clinical outcome. If many diseased subjects are missed, they undesirably go without treatment. On the other extreme, if the screening service is overly conservative and refers many normal subjects to further clinical assessment, it unnecessarily burdens a significant proportion of the subjects and keeps trained clinicians occupied with medically irrelevant cases, and eventually loses credibility. Thus the principal challenge here lies in ensuring high level of screening accuracy without involving ophthalmologists. In response, we propose an accurate yet low-cost screening service, which, by leveraging machine learning tools, requires only limited inputs from personnel with relatively less training. 

Here, one important issue arises in choosing an appropriate imaging modality. As development of such screening tool has received rather limited attention in the context of eyecare, existing tools do not provide reliable guidance in making a choice. One recently suggested screening tool is based on fundus photography (FP) \cite{acharya2016automated}. However, in this paper, we make use of FAF imaging in view of various extensive investigations that compared GA assessment based on FP images and that based on FAF images, and indicated higher consistency and reproducibility in case of the latter modality \cite{iovs2007comparing}\cite{retina2012comparison}\cite{hadi2013comparison}. More generally, FAF imaging has been found to facilitate detection of abnormalities beyond those detected based on funduscopic exam, fluorescein angiography, or optical coherence tomography, and can be used to elucidate disease pathogenesis, form genotype-phenotype correlations, diagnose and monitor disease \cite{FAF_clinical}. Further, FAF imaging has been reported to lead to better understanding in disease-specific circumstances, including choroidal neovascularization (CNV) in AMD \cite{FAF_CNV}. In short, FAF imaging has been understood to provide more reliable quantification compared to FP, but utilized principally in the context of diagnosis. In the present paper, we propose FAF imaging as the basis for screening of diseased eyes, in order to take advantage of its superiority. In addition, the aforesaid FP-based tool performed empirical mode decomposition, estimated entropy and energy of each decomposed image, and used those as features to distinguish between FP images of healthy and diseased eyes \cite{acharya2016automated}. However, such features are global, and do not capture local information about spatial variation, which, as alluded earlier, disease-specific studies have found clinically relevant. Accordingly, for the proposed screening tool based on FAF imaging, we adopt a different set of clinically inspired features. In this connection, we hypothesize that the previously mentioned sectoral statistics of FAF images, based on which indicators for specific pathologies have already been suggested \cite{Retina2013, schmitz2008evaluation, jolly2016novel, zola2018evolution}, also possesses general information about diseased conditions, and hence remain relevant even for screening.
Specifically, given a FAF image, we propose to center an early treatment of diabetic retinopathy study (ETDRS) grid \cite{early1991early}, a well accepted tool in ophthalmic studies, at the fovea, and compute mean and standard deviation of pixel intensity within each sector \cite{Retina2013}. Collecting such sectoral statistics in a feature vector, a classifier is then trained to distinguish FAF images of healthy eyes from the diseased ones in such feature space.

Next, we turn to choosing a classifier for prediagnosis screening. Here, we also keep in mind that a patient often requires periodic monitoring of disease progression, post diagnosis \cite{schmitz2009fundus}. Accordingly, to keep the system cost low, we seek such a technological platform for the desired screening classifier that also potentially supports an efficient system for monitoring disease progression. In this regard, a recent work on retinal health screening adopted features based on empirical mode decomposition, and employed support vector machine (SVM) classifier  \cite{acharya2016automated}. Such a classifier in general attempts to learn a decision boundary between diseased and healthy classes in the feature space from class labels \cite{cortes1995support}, and is potentially attractive even for disease progress monitoring. Specifically, upon diagnosing of a specific disease, one could note whether a patient's features gradually move further from (or closer to) the decision boundary over several visits, and possibly infer improvement in (or worsening of) disease condition. In other words, SVM classifier could supply the desired common platform alluded earlier, provided that it achieves suitably high screening accuracy. Accordingly, the main focus of this paper lies in establishing suitability of SVM for ophthalmic disease screening using the proposed sectoral features.

At this point, it is worthwhile to note that SVM classifiers can be implemented using different transformation kernels, the linear function and radial basis function (RBF) being the most ubiquitous ones \cite{LIBSVM}. Accordingly, we perform a comparative study between linear SVM and RBF-SVM in terms of screening performance, and report satisfactory performance of both the classifiers. However, we recommend the use of the latter, which exhibits a slight empirical superiority. In this connection, note that the aforesaid recent screening method also used RBF-SVM (albeit in association with a different set of features) but made no comparison with linear SVM \cite{acharya2016automated}. Finally, to ensure improved fairness in evaluating the classifier performance, we adopt Monte Carlo cross validation (MCCV) rather than the more ubiquitous K-fold cross validation. 

The rest of the paper is organized as follows. FAF image acquisition, ETDRS grid placement and the computation of statistics, SVM classifiers and performance evaluation criteria are described in Section \ref{sec:MM}. In Section \ref{sec: RES}, experimental results on screening performance and statistical analysis are presented. Additional observations related to performance of other classifiers and possible monitoring of disease progression using SVM
are made in Section \ref{sec: OtherClass} and Section \ref{sec: ProgMon}. The future course, covering potential extensions and practical implementation, is outlined in Section \ref{sec:future}.


\section{MATERIALS AND METHODS}
\label{sec:MM}

\begin{figure}[t!]
\centering
    \includegraphics[scale=0.3]{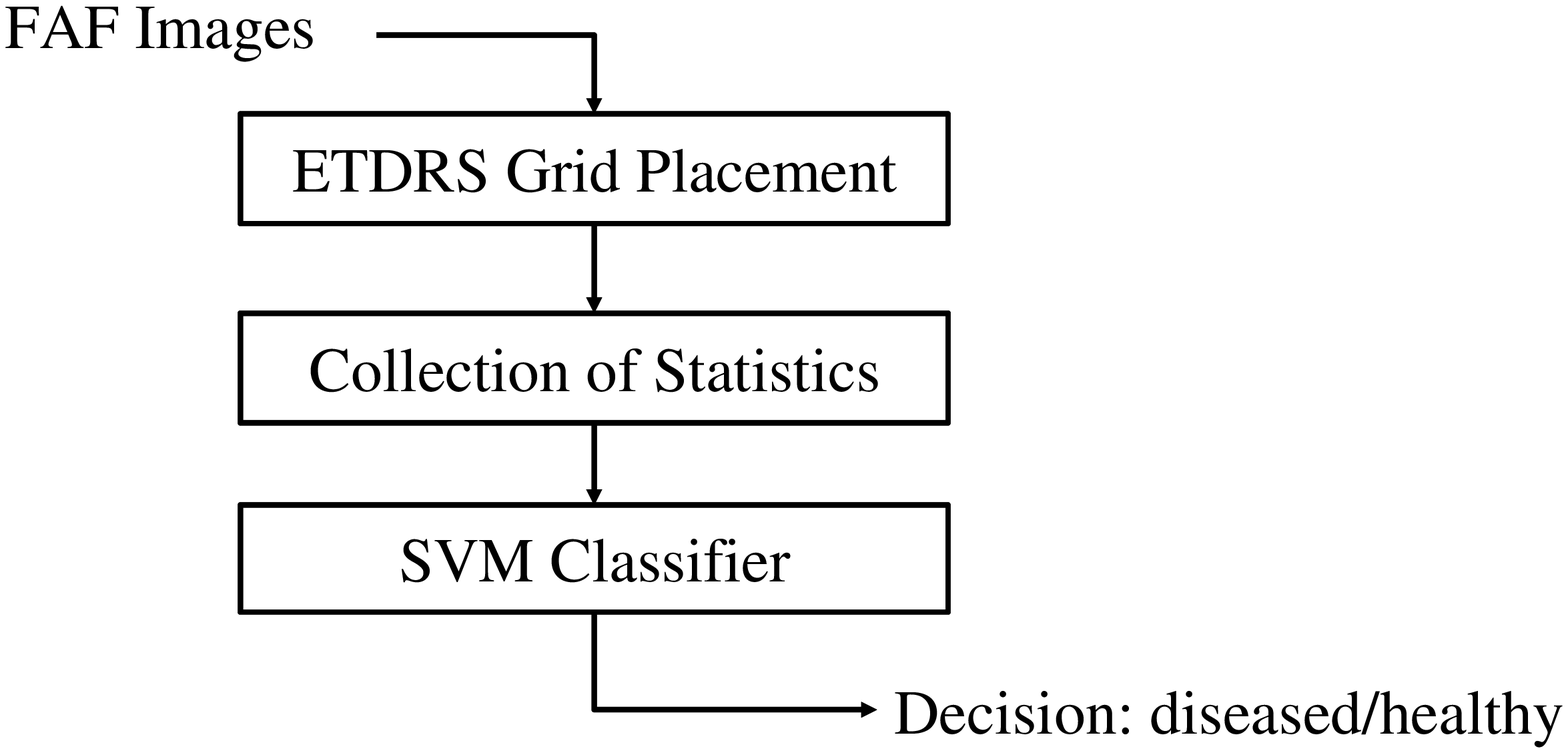}\par

\caption{Schematic diagram of proposed screening method.}
\label{fig:flow}
\end{figure}

In this section, we propose and describe a screening methodology, outlined in Figure ~\ref{fig:flow}, where the ETDRS grid is placed on the input FAF image, and an SVM classier attempts to detect occurence of disease based on sectorial statistics. First, we detail the method of acquisition of FAF images.

\subsection{Fundus autoflourescence image acquisition}
\label{sec: FAF}

This is a retrospective study conducted at tertiary eye centers of L V Prasad Eye Institute (LVPEI) in South India. A total of 140 subjects underwent fundus autoflourescence examination where 61 are healthy and rest are diagnosed to have ophthalmic diseases. Specifically, of the 79 subjects, 21 of them are diagnosed with CSCR, 14 with choroidal neovascular membranes (CNVM), and 44 with STGD. The procedures performed in this study conformed to the tenets of Declaration of Helsinki. Respective Institutional Review Boards approved the study protocol. Written informed consent was obtained from all the study participants. The Heidelberg Retina Angiograph 2 (HRA2; Heidelberg Engineering, Heidelberg, Germany) was used to acquire the FAF images. Standard protocol for image acquisition involved excitation at 488 nm with an optically pumped solid-state laser and emission detected above 500 nm with a barrier filter. Typically, between 9 and 15 images were averaged to acquire a mean image. Either the automatic real-time mode or the mean image mode was used for all FAF image acquisition. This protocol allowed for an amplified signal with reduced noise for each image.

\subsection{ETDRS-based statistics as feature set}
\label{sec: Stats}

\begin{figure}[t!]
\centering
    \includegraphics[width=0.75\columnwidth]{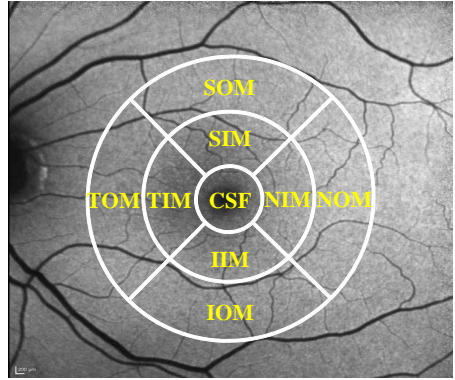}\par 

\caption{FAF image with ETDRS grid.}
\label{fig:ETDRS}
\end{figure}

In the current work, the ETDRS grid was used for computing features from each FAF image with the goal of classifying it as either `healthy' or `diseased'. The ETDRS grid, positioned by an optometrist (rather than an ophthalmologist) at the centre of the fovea as shown in Figure ~\ref{fig:ETDRS}, has nine segments. The inner most ring, known as central subfield (CSF), is sorrounded by pericentral ring consisting of temporal inner macula (TIM), inferior inner macula (IIM), nasal inner macula (NIM) and superior inner macular (SIM) regions. Further, the peripheral ring in ETDRS grid consists of temporal outer macula (TOM), inferior outer macula (IOM), nasal outer macula (NOM) and superior outer macular (SOM) regions. In each of the nine segments, the mean pixel intensity and the corresponding standard deviation were computed. Such statistics from all nine segments formed a feature vector of length 18 representing each FAF image. These features were used to distinguish between diseased and healthy eyes.

\begin{figure}[t!]
\centering
   \includegraphics[width=0.24\columnwidth]{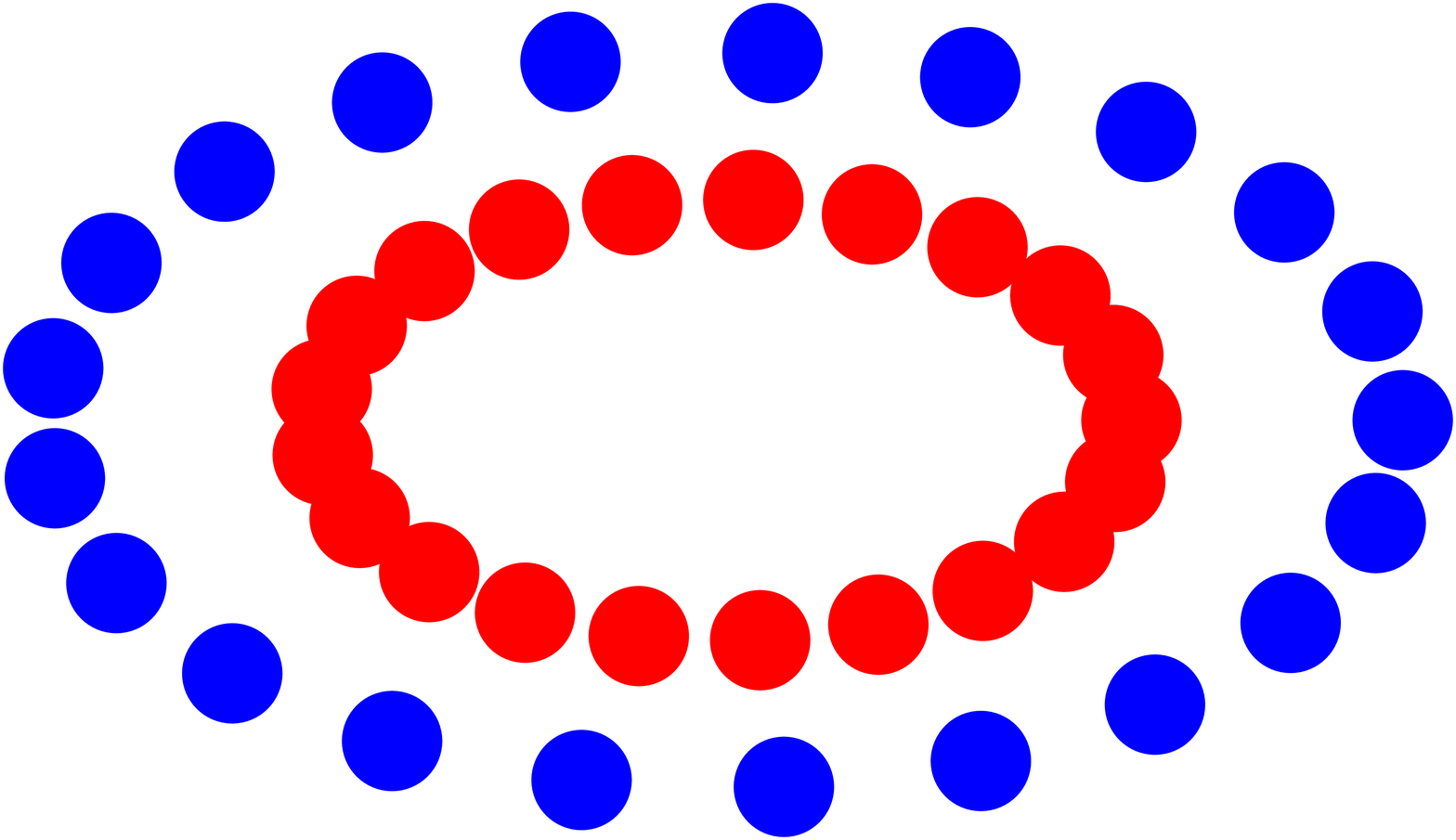}
   \includegraphics[width=0.34\columnwidth]{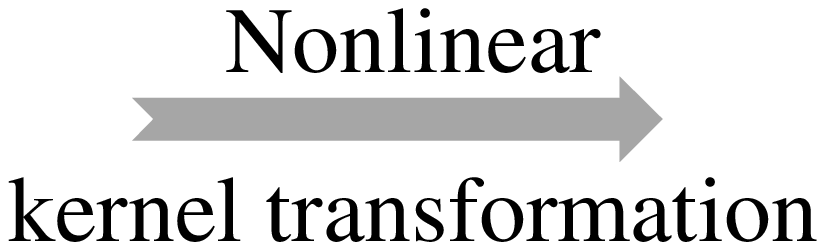}
   \includegraphics[width=0.39\columnwidth]{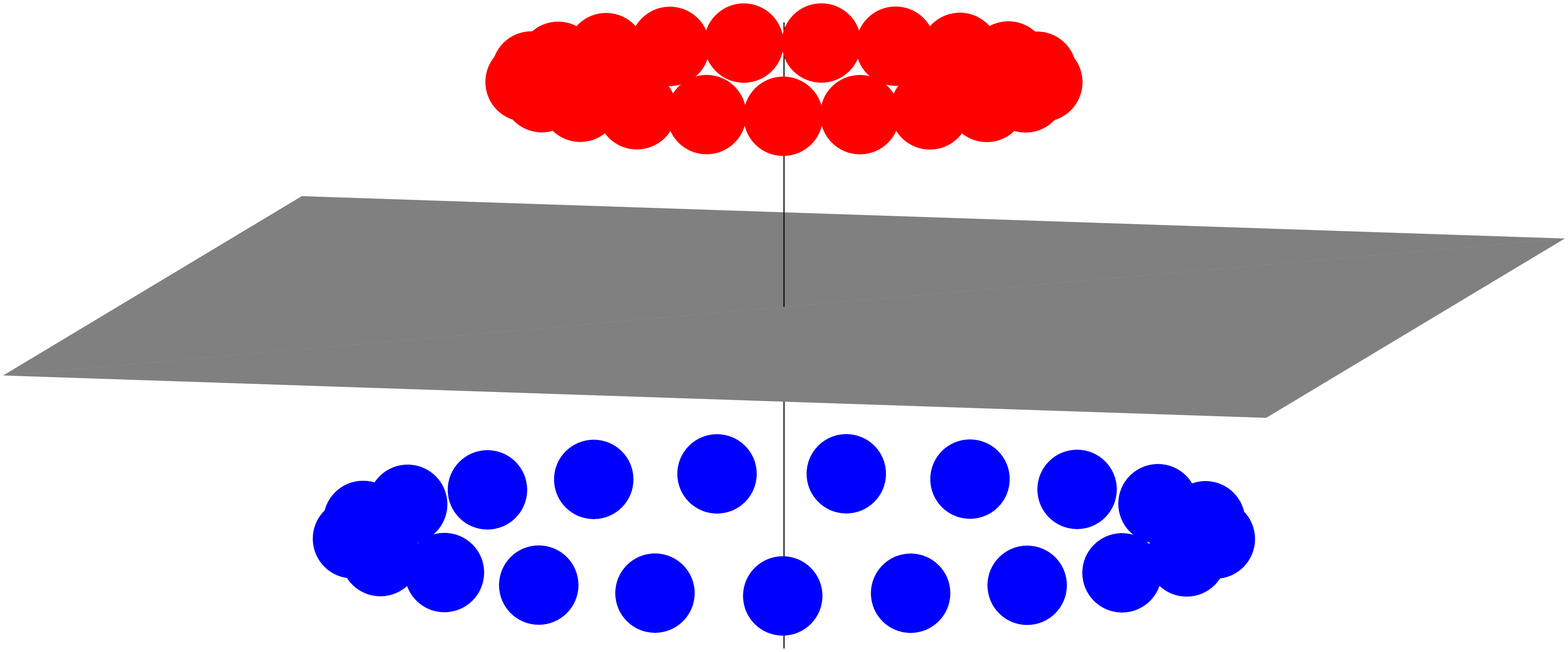} \\

\caption{SVM with nonlinear kernel: data points belonging to each class indicated by unique color; potential linear separability upon nonlinear kernel transformation.}
\label{fig:svmfig}
\end{figure}


\subsection{Support vector machine (SVM) classifier}
\label{sec: SVM}


The support vector machine (SVM) classifier provides a supervised learning model \cite{cortes1995support}, which upon training, obtains a hyperplanar decision boundary maximally separating two classes. More generally, to handle data that are nonlinearly separable, SVM can incorporate suitable kernel that transforms the original feature space into a higher-dimensional space, where the transformed features become linearly separable (See Figure \ref{fig:svmfig} for an explanatory illustration). Formally, the decision hyperplane $w^T \phi(x) + b = 0$ (where $\phi(x)$ denotes a point vector in the transformed feature space, $\phi$ the kernel function, $w$ the weight vector and $b$ the bias) is chosen to maximize overall separation. One equivalently maximizes
$$\mbox{ {$\sum_{i=1}^{n} \alpha_i$ - 
$0.5 \sum_{i=1}^{n}\sum_{j=1}^{n} \alpha_i \alpha_j y_i y_j K(x_i,x_j)$ }}
$$
over $\alpha_i\geq 0$, $i=1,2,...,n $, subject to $ \sum_{i=1}^{n} \alpha_i y_i = 0$. Here, for each data index $i$, $x_i$ and $y_i$, respectively, indicate the feature vector and the corresponding class label ($+1$ for diseased; $-1$ for healthy), and $K$ denotes the kernel function, $n$ the size of the dataset, $\alpha_i$ the Lagrange multiplier. We compare classification performance of two variants of SVM, one with linear kernel (where $K(x_i,x_j) = x_i^T x_j$),and the other with radial basis function (where $K(x_{i},x_j) = \exp{\frac{\left | {x_i -x_j} \right |^2}{2\sigma^2}}$), an ubiquitous nonlinear kernel.

\begin{figure}[t!]
\centering
   \includegraphics[width=1\columnwidth]{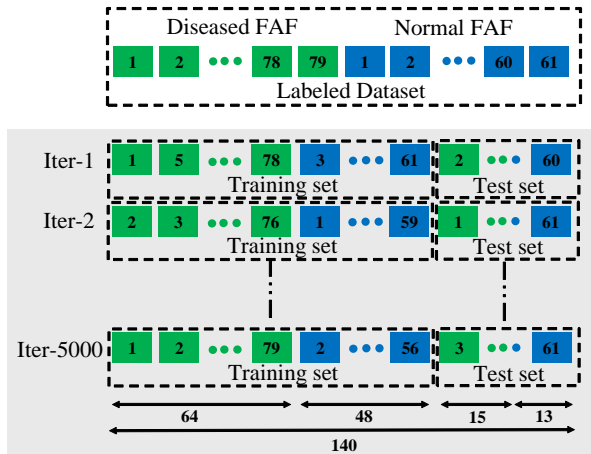}

\caption{Illustration of MCCV method.}
\label{fig:MCCV}
\end{figure}


\subsection{Performance evaluation}
\label{sec: Eval}

To evaluate classification performance, the SVM parameters, namely, weight vector $w$ and bias $b$, are optimized on a subset of data (training subset) and are validated against the complementary (test) subset. More accurately, the dataset is partitioned into training and test subsets such that (i) the ratio of their sizes, called training-to-test (split) ratio, approximately equals a preassigned fraction, and (ii) the proportion of healthy (as well as diseased) images represented in those subsets also approximately equals the same fraction. In general, the performance of the classifier depends on the subjective choice of the partition. To avoid such subjectivity in performance analysis, one customarily uses Monte Carlo cross validation (MCCV) ~\cite{xu2001monte}, where, as shown in Figure \ref{fig:MCCV}, the dataset is randomly partitioned a large number (5000) of times (iterations) keeping the split ratio as well as the aforementioned proportions constant. For each iteration, the SVM parameters are optimized over the training subset, and the mean training as well as test accuracy and corresponding standard deviation are recorded. Noting that the training accuracy indicates the classification performance for seen data, while the test accuracy indicates that for unseen data, classifiers with high average test accuracy assumes practical significance. Further, low standard deviation, indicating low performance variability over random partitions and hence signifying robustness, is desirable. Additionally, we also compute the average confusion matrix over 5000 iterations, which provides class-conditional detection probability for each of healthy and diseased classes. Finally, we train and evaluate classifiers for different split ratios.

\begin{figure*}[t!]
\centering
    \begin{tabular}{ccc}
    \includegraphics[width=4.5cm]{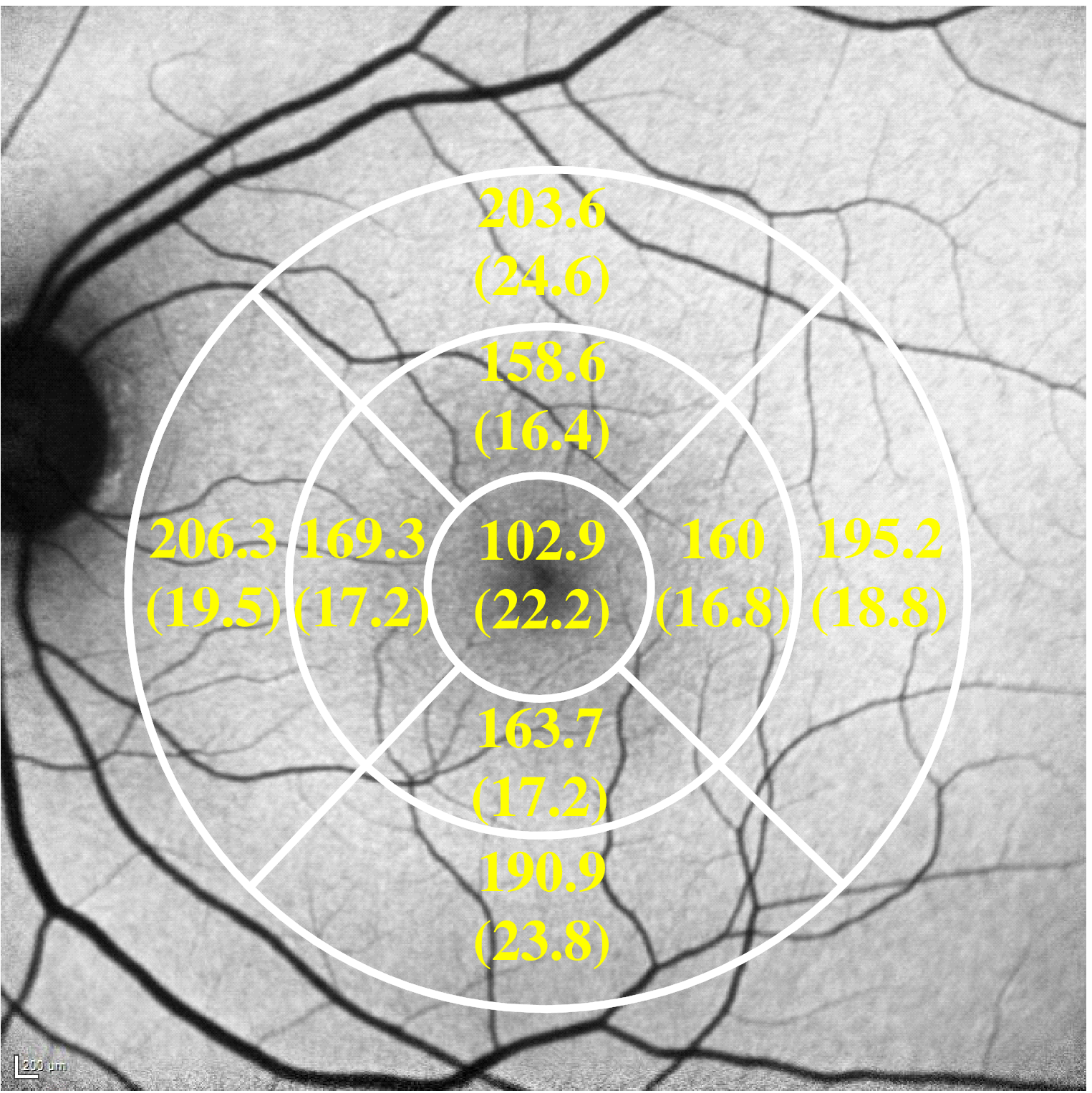}& 
    \includegraphics[width=4.5cm]{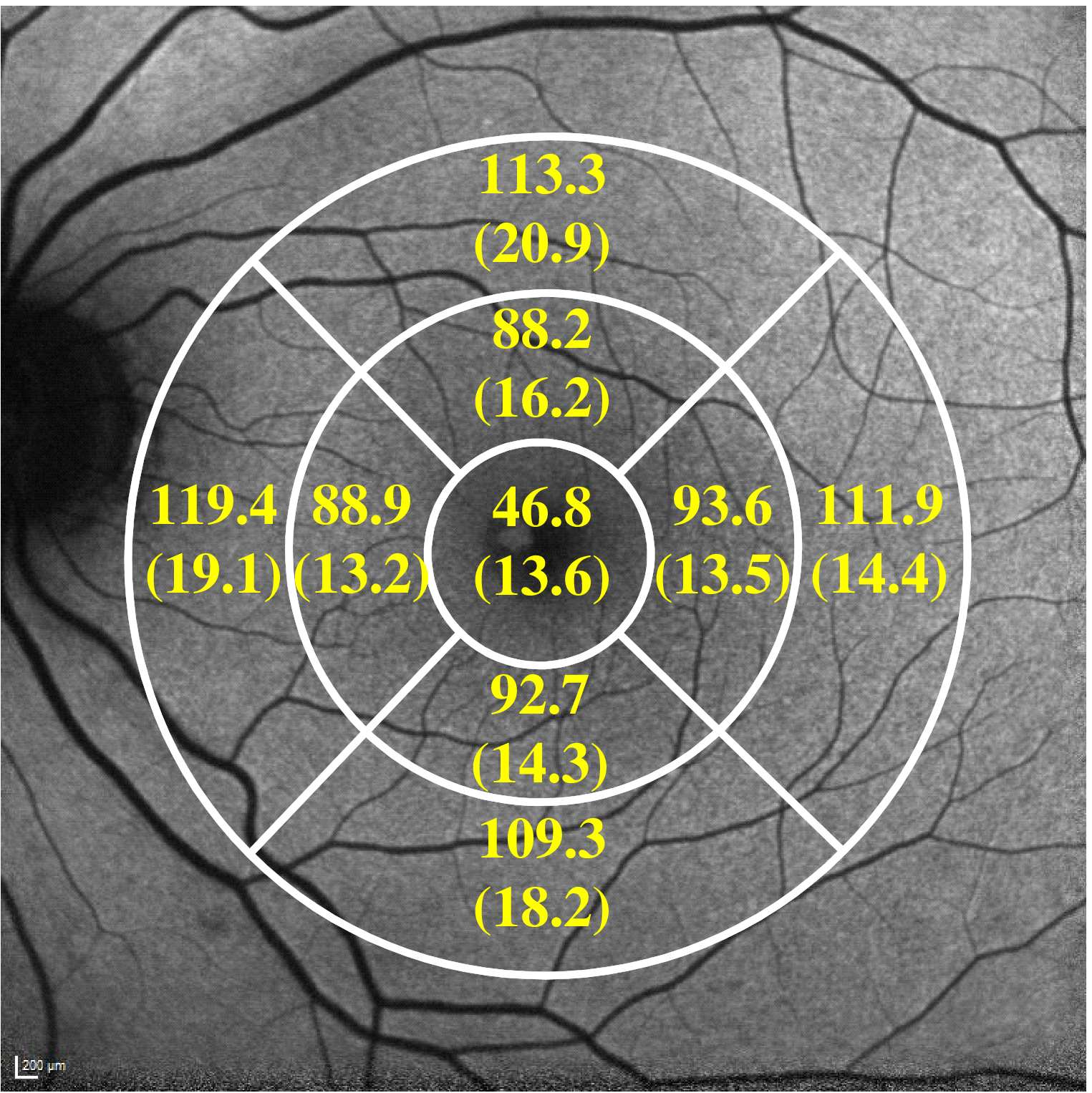}& 
    \includegraphics[width=4.5cm]{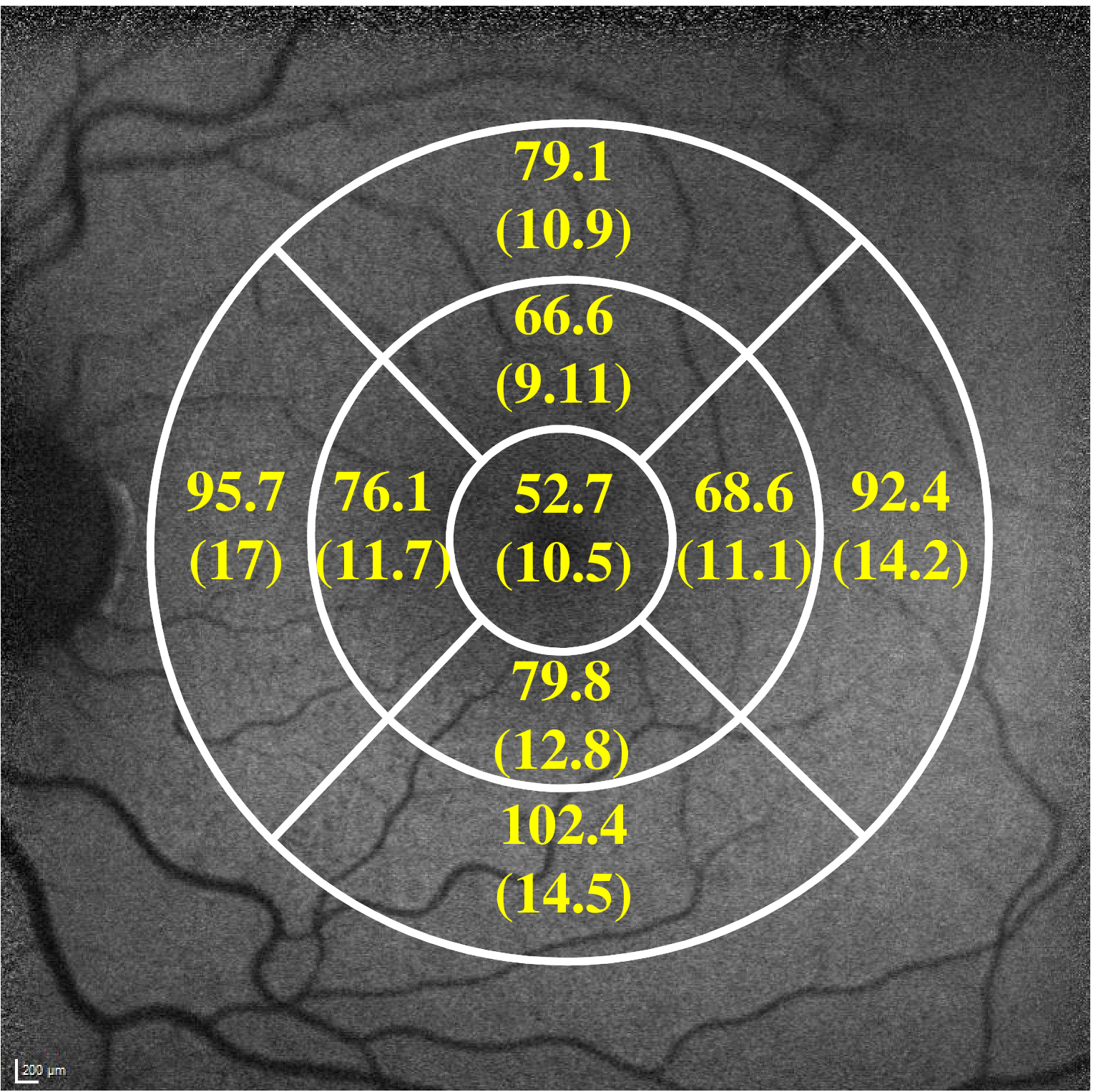}\\
    &(a)&\\
    \includegraphics[width=4.5cm]{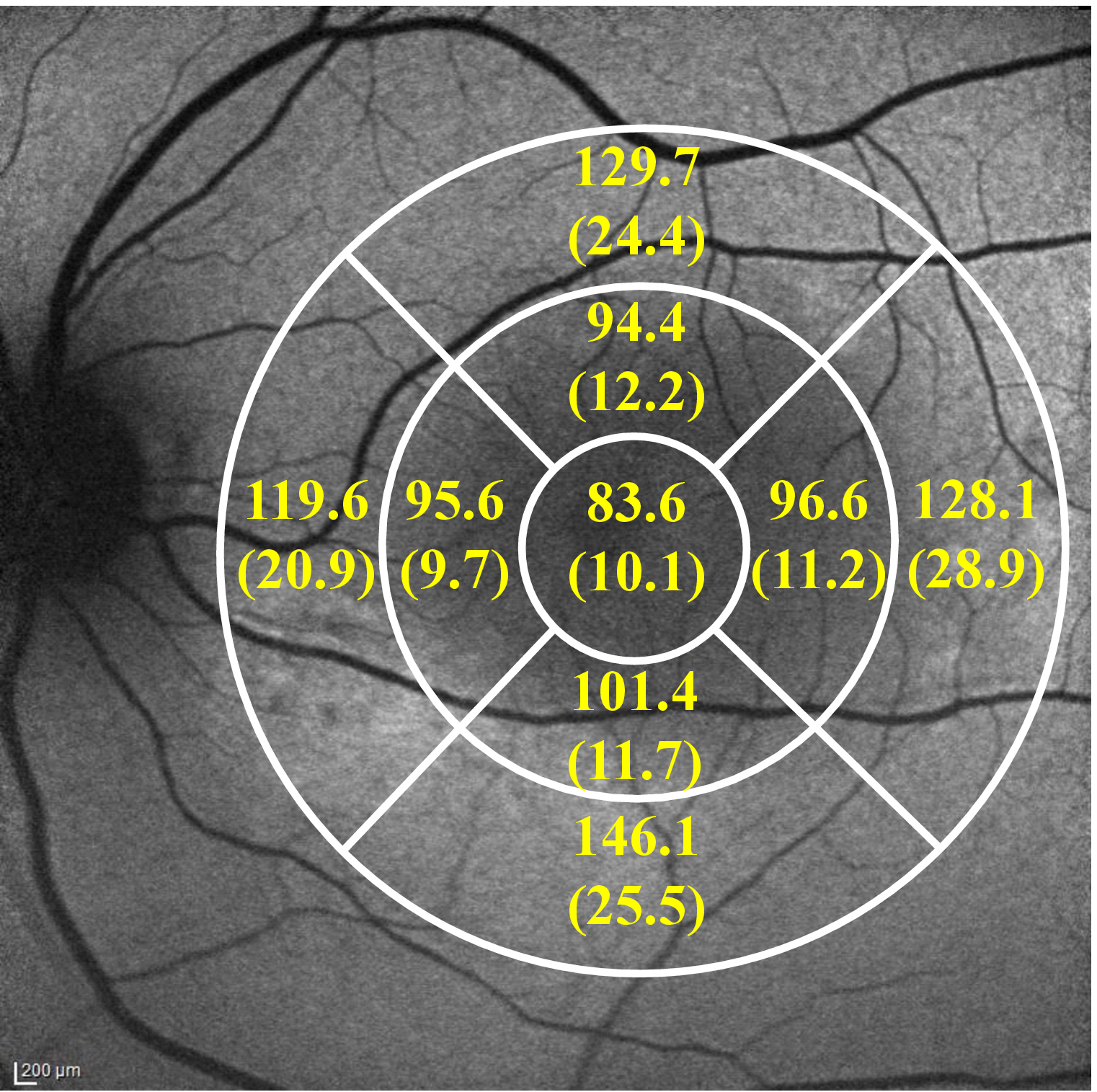}& 
    \includegraphics[width=4.5cm]{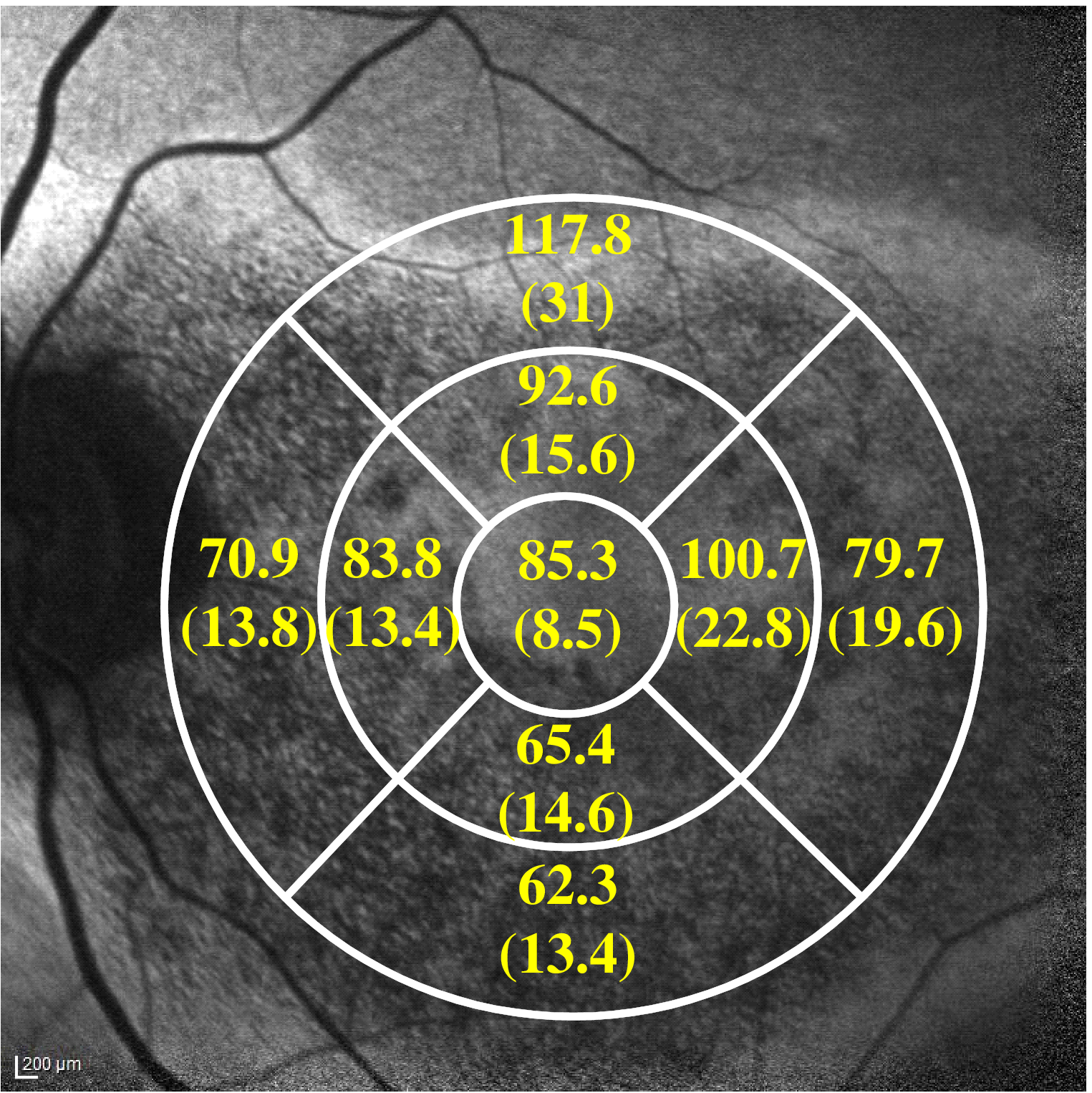}& 
    \includegraphics[width=4.5cm]{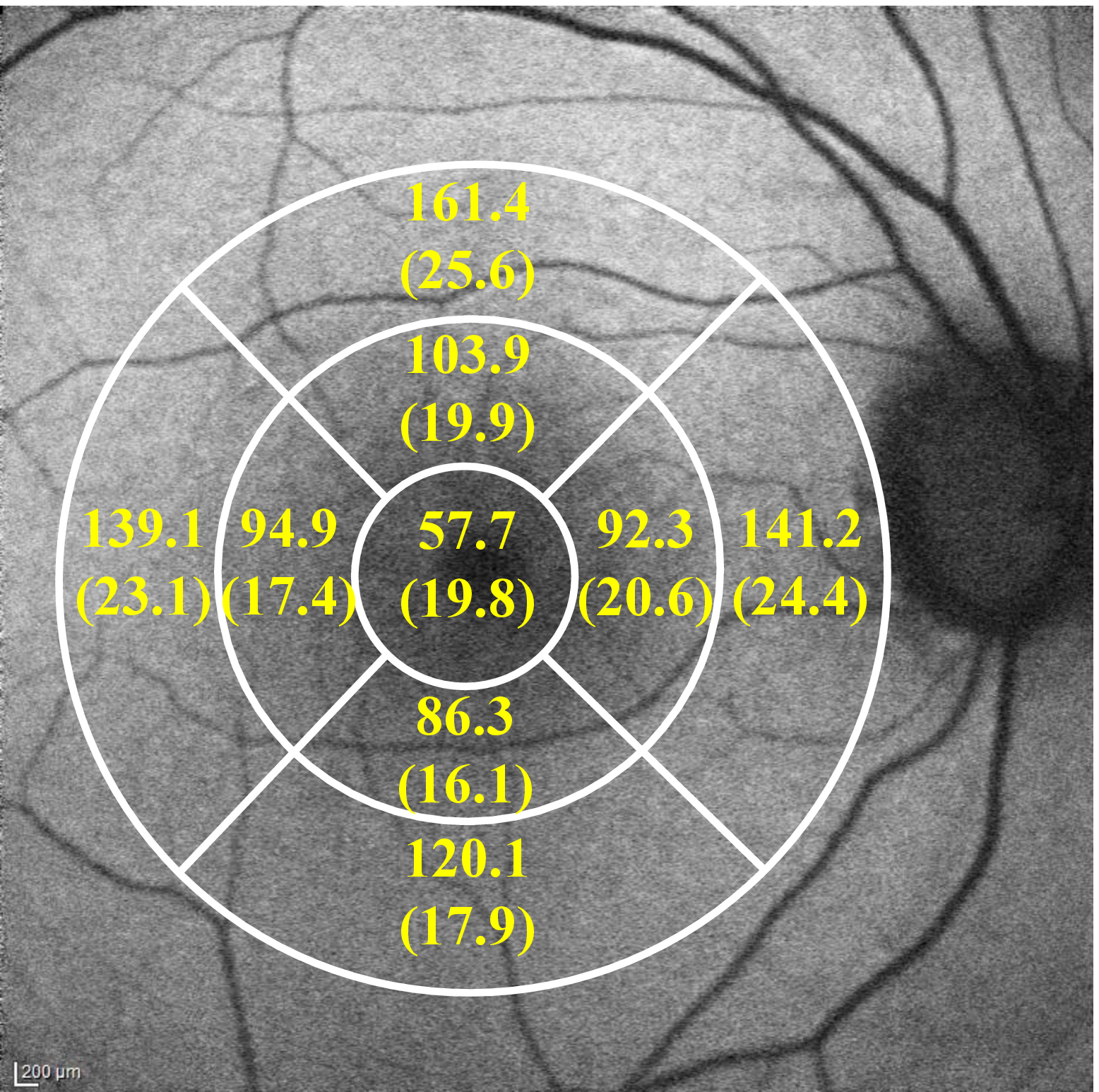}\\
    & (b)&\\

 \end{tabular}  
\caption{ETDRS grid placement on and corresponding sectorial statistics (mean with standard deviation in parenthesis) for representative images of (a) healthy eyes and (b) diseased eyes.}
\label{fig:gridstats}
\end{figure*}

\section{EXPERIMENTAL RESULTS}
\label{sec: RES}

As described earlier, placing ETDRS grid on each FAF image, we computed mean and standard deviation of pixel values corresponding to each of the nine sectors (Figure \ref{fig:ETDRS}), and thus formed a feature vector of length 18. In Figure \ref{fig:gridstats}, such sectoral features are illustrated for example FAF images of both healthy and diseased eyes. As mentioned earlier, two variants of SVM classifiers, linear SVM and SVM with RBF kernel (RBF-SVM), were considered. In each case, training and test accuracy values were recorded over a large number (5000) of random partitions for each of the various training-to-test split ratios chosen between 10:90 and 90:10, and 
the average values (with standard deviation values in parenthesis) are tabulated in Table~\ref{table:res}.

\begin{table*}[t!]
\centering
\caption{Performance of rival SVM clasifiers for different splits (superior value between linear SVM and RBF-SVM boldfaced) }
\label{table:res}
\begin{tabular}{cccccccc}
\toprule
        &            \multicolumn{2}{c}{Linear SVM } &  \multicolumn{3}{c}{RBF-SVM } &  \multicolumn{2}{c}{Gain (\%)}\\
\cmidrule(lr){2-3}
\cmidrule(lr){4-6}
\cmidrule(lr){7-8}
  Split   & Train. Acc. & Test Acc. & & Train. Acc.  & Test Acc. & Train. Acc. & Test Acc. \\
  ratio   & $a$ ($b$) & $c$ ($d$) &  SF & $e$ ($f$) & $g$ ($h$) &$\frac{e-a}{a}$ ($\frac{b-f}{b}$)&   $\frac{g-c}{c}$ ($\frac{d-h}{d}$) \\
\cmidrule(lr){1-1} \cmidrule(lr){2-2} \cmidrule(lr){3-3}
\cmidrule(lr){4-4} \cmidrule(lr){5-5} \cmidrule(lr){6-6}
\cmidrule(lr){7-7} \cmidrule(lr){8-8} 


{10:90}   & 91.27 (7.01) &  75.03 (7.05)   &  2.85 & \textbf{99.38 (2.13)} & \textbf{78.97 (5.30)} & 8.16 (69.61) & 4.99 (24.82)\\


{20:80}   & 91.51 (4.82)  &  82.26 (4.78)   &  2.80 & \textbf{98.81 (2.09)}  & \textbf{84.32 (3.96)} & 7.39 (56.65) & 2.44 (17.15)\\


{30:70}   & 92.04 (3.66) &  85.56 (3.81) & 3.20 & \textbf{97.88 (2.17)} & \textbf{86.66 (3.48)} & 5.97 (40.71)  &  1.27 (8.66) \\


{40:60}   & 92.26 (2.91) &  87.31 (3.46) &   2.70 & \textbf{98.86 (1.38)}  & \textbf{88.11 (3.35)} & 6.67 (52.27) &  0.91 (3.18) \\

{50:50}   & 92.24 (2.44) & 88.40 \textbf{(3.43)} &  3.00 & \textbf{98.86 (1.38)} & \textbf{88.89 (3.43)} & 6.69 (43.44) &  0.55 (0.00) \\


{60:40}   & 92.27 (2.01) & 89.02 (3.68) & 2.90  & \textbf{98.46 (1.20)}  & \textbf{89.58 (3.47)} & 6.28 (40.29) &  0.62 (5.71) \\


{70:30}   & 92.31 (1.64) &89.53 \textbf{(4.14)} &  2.80   &   \textbf{98.62 (0.99)} & \textbf{90.10} (4.16)   & 6.39 (39.63)  & 0.63 (-0.48) \\


{80:20}   & 92.26 (1.30) & 89.88 \textbf{(5.12)} &  2.75   & \textbf{98.65 (0.90)}   & \textbf{90.55} (5.24) & 6.47 (30.77) &  0.75 (-2.34) \\


{90:10}   & 92.25 (1.01) & 89.60 \textbf{(7.16)} & 2.65 & \textbf{98.86 (0.71)} & \textbf{90.83} (7.21) &  6.68 (29.70) &  1.35 (-0.70) \\

    
\bottomrule
\end{tabular}

\end{table*}

Clearly, in case of linear SVM, the average training as well as test accuracy level tends to increase with increasing training-to-test ratio but for a few exceptions. In other words, as expected, the said classifier tends to learn better with increased availability of training data. Turning to the RBF-SVM classifier, for each split ratio we chose the scale factor (SF) that maximizes test accuracy. The resulting average test accuracy, which still largely follows the aforementioned increasing trend, slightly improves over that obtained using the linear SVM for each split ratio. This indicates a moderate degree of nonlinearity inherent in the underlying problem. Despite only a slight gain in test accuracy, the gain in average training accuracy is significantly higher, indicating that the RBF kernel is closely modeling certain nonlinear aspects of the training data that do not generalize well. In fact, a similar conclusion can also be drawn by considering standard deviation values as follows. Notice that the standard deviation in training accuracy decreases with increasing split ratio in both cases of linear SVM and RBF-SVM; however, the values are significantly lower for the latter classifier, indicating better modeling. Yet, turning to test accuracy, the standard deviation is lower in case of RBF-SVM for lower values of split ratio and in case of linear SVM for higher values of split ratio. This phenomenon does not indicate improved generalization. Further, the standard deviation in either case first decreases and then increases with increasing split ratio. Yet a subtle difference can be discerned. While the test accuracy is most reliable (i.e., with the lowest standard deviation) for an evenly split training-to-test ratio in case of linear SVM (also observed elsewhere \cite{milk}), highest reliability is observed at the skewed split ratio of 40:60 in case of RBF-SVM. 

\begin{figure}
\centering
 \includegraphics[scale = 0.15]{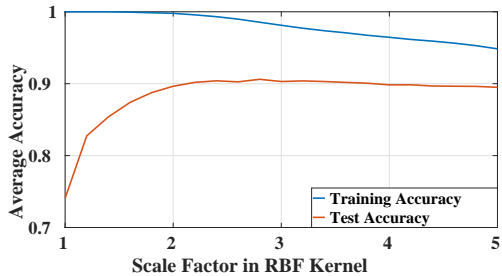}
\caption{Average training and test accuracy of RBF-SVM classifier over MCCV with variation in scale factor (80:20 split ratio).}
\label{fig:ScalFac}

\end{figure}

Having reported performance variation over the entire range of split ratios, we next delve deeper into case of 80:20 split, a general recommendation for various practical applications \cite{macek2008}. First, we studied the performance variation with varying SF. In particular, the SF was varied between 1 and 5, and corresponding training and test accuracy levels over 5000 MCCV iterations were recorded. Notably, as seen in the Fig. \ref{fig:ScalFac}, the training accuracy decreases with increase in SF, whereas the test accuracy increases with increase in SF upto a maximum value of 2.75. For larger values of SF, the test accuracy decreases so slowly that it may essentially be taken as unchanged. Thus, we take RBF-SVM with SF$=2.75$ as the classifier competing with the linear SVM. In particular, we next study class-conditional performance of these rival classifiers at hand. Practical significance of this study arises from the fact that mistakenly classifying a diseased eye as normal leads to the severe consequence of denial of treatment to a bona fide patient, whereas declaring a normal eye as diseased leads to the comparatively less severe outcome of wasted time and resources. Referring to Table \ref{table:conf}, RBF-SVM records an average probability of 91.61\% in correctly detecting the disease class whereas it is only 89.47\% using linear SVM, with respective standard deviations 7.09\% and 7.33\%. In contrast, conditioned on the healthy class, the healthy class was correctly detected with an average probability of 92.25\% by linear SVM and 88.59\% by RBF-SVM, with respective standard deviations 7.11\% and 8.82\%. Thus, given the healthy class, linear SVM slightly outperforms RBF-SVM in terms of both accuracy and reliability. Notice that the advantage of RBF-SVM over linear SVM given the diseased class is comparable to that of linear SVM over RBF-SVM given the healthy class. However, since making an error incurs higher practical cost given the diseased class than that given the healthy class, one should prefer RBF-SVM over linear SVM as a screening tool. In summary, based on our experiments, we conclude that SVM classifiers of FAF images, attaining close to 90\% accuracy, provide attractive screening tools for ophthalmic diseases. Among those, RBF-SVM appears to provide slight advantage over linear SVM in terms of both overall accuracy levels and practical class-conditional costs.

\begin{table}[t!]

\centering
\caption{Test confusion matrices (\%) of linear SVM and RBF-SVM for split ratio 80:20 for original screening dataset}
\label{table:conf}
\begin{threeparttable}
\resizebox{\columnwidth}{!}{
\begin{tabular}{cccc}

 \toprule
& \diagbox[width=2.7cm]{Predicted}{Actual}  & Diseased & Healthy \\
    \hline
Linear SVM & Diseased & 89.47 (7.33) & 7.75 (7.11)  \\
 & Healthy & 10.53 (7.33) & 92.25 (7.11)   \\
 \hline 
RBF-SVM & Diseased & 91.61 (7.09) & 11.41 (8.82) \\
 & Healthy &  8.39  (7.09) &  88.59 (8.82)   \\




\bottomrule
\end{tabular}}

\end{threeparttable}

\end{table}

\begin{table}[t!]
\centering
\caption{Test confusion matrix (\%) of RBF-SVM for split ratio 80:20 for the reduced dataset (diseased class represented by STGD only)}
\label{table:AMDconf}
\begin{threeparttable}
\resizebox{0.85\columnwidth}{!}{
\begin{tabular}{ccc}
\toprule

\diagbox[width=2.7cm]{Predicted}{Actual}  & Diseased & Healthy \\
\cline{1-3}

Diseased & 99.35 (2.65) &  0.98 (2.61) \\
 Healthy & 0.65 (2.65) & 99.02 (2.61) \\

\bottomrule
\end{tabular}}

\end{threeparttable}

\end{table}

Finally, it is worthwhile to demonstrate that the screening problem poses harder challenges compared to detection of a specific disease. Recall that our screening dataset consists of FAF images of healthy eyes as well as eyes affected by highly prevalent diseases STGD, CNVM and CSCR, between which we did not distinguish so far. At this point, consider the disease detection problem instead, where one checks whether a particular disease (say, stargardt) is present. Specifically, we now decide between healthy versus stargardt-affected eyes, and note for RBF-SVM the marked improvement in the corresponding class-conditional accuracy levels (99.35\% and 99.02\%, respectively) given in Table \ref{table:AMDconf} over those (91.61\% and 88.59\%, respectively; see Table \ref{table:conf}) in case of disease-agnostic screening, indicating the relative difficulty posed by the screening problem.

\balance


\section{Additional Performance Comparison}
\label{sec: OtherClass}

\begin{figure}[t!]
\centering
 \includegraphics[scale = 0.215]{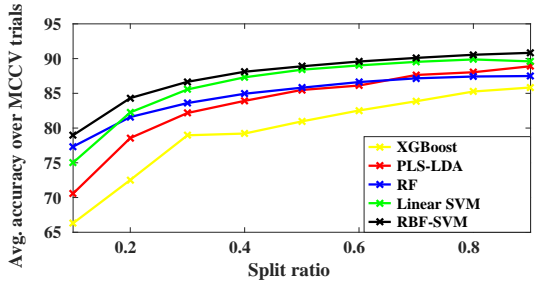}
\caption{Superiority of RBF-SVM over other classifiers across varying split ratio.}
\label{fig:AllClass}

\end{figure}

So far, our focus has  remained on the development of an accurate SVM-based automated screening tool that detects possible presence of ophthalmic disease. In this context, one would wonder how the proposed SVM-based classifier compares in terms of performance with other well-known alternatives that have proven efficacious in related contexts. Accordingly, we considered partial least squares trained linear discriminant analysis (PLS-LDA\footnote{PLS-LDA performs supervised dimensionality reduction using partial least squares regression, and thence learns a hyperplane separating two classes such that intra-class variance is minimized and inter-class variance is maximized.}) \cite{LDA} (effective in diagnostic quality assessment of fundus photographs \cite{DQAembc19}), extreme gradient boosting of decision trees (XGBoost) \cite{xgboost} (effective in classifying retinal cysts \cite{SRFPEDembc19}), and random forest (RF\footnote{Both XGBoost and RF methods employ ensemble learning on multiple decision trees, albeit using gradient boosting via Newton-Raphson approximations and bootstrap aggregation (bagging), respectively.}) classifiers \cite{RF} (effective in detecting diabeties \cite{MLtools}, diabetic retinopathy and age-relate macular degeneration based on fundus images \cite{ICIP2015}). 
For the present screening problem, we plotted in Fig. \ref{fig:AllClass} the average test accuracy obtained respectively by the SVM variants and other classifiers at hand against variation in training-to-test split ratio. We observe that the proposed SVM-RBF classifier outperforms the rivals across split ratios. In fact, both SVM classifiers generally prove superior to the rest of the classifiers under consideration. As an aside, PLS-LDA, while inferior to the SVM classifiers, performs superior to RF as well as XGBoost at 80:20 split. Here, hyperparameters were tuned (details omitted for the sake of brevity and continuity) for each of PLS-LDA, XGBoost and RF classifiers as well, so as to ensure fairness of comparison with the tuned SVM classifiers.


\section{Possible Progress Monitoring}
\label{sec: ProgMon}

\begin{figure*}[t!]
\centering
\begin{tabular}{cc}

  \includegraphics[width=1\columnwidth]{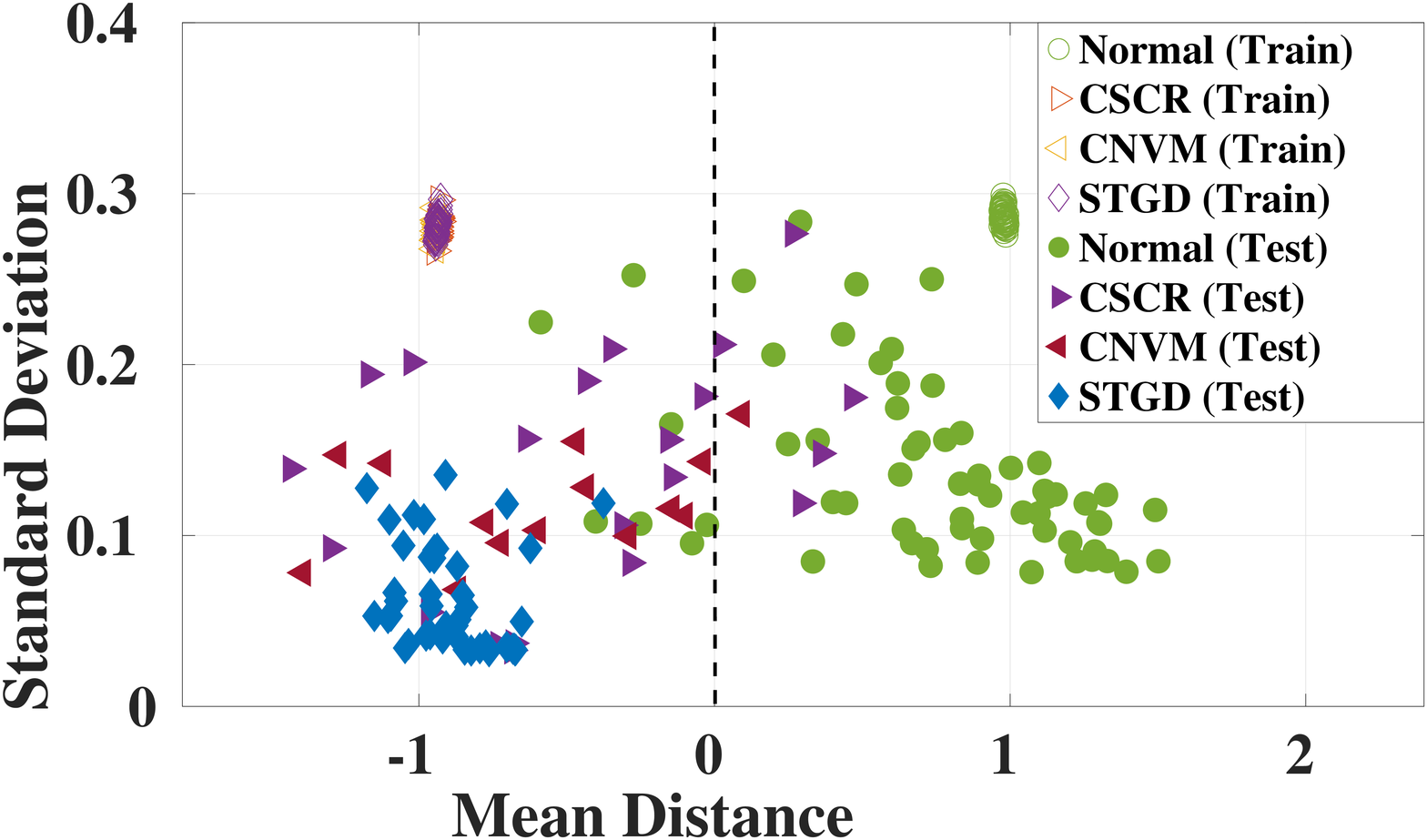} &
 \includegraphics[width=1\columnwidth]{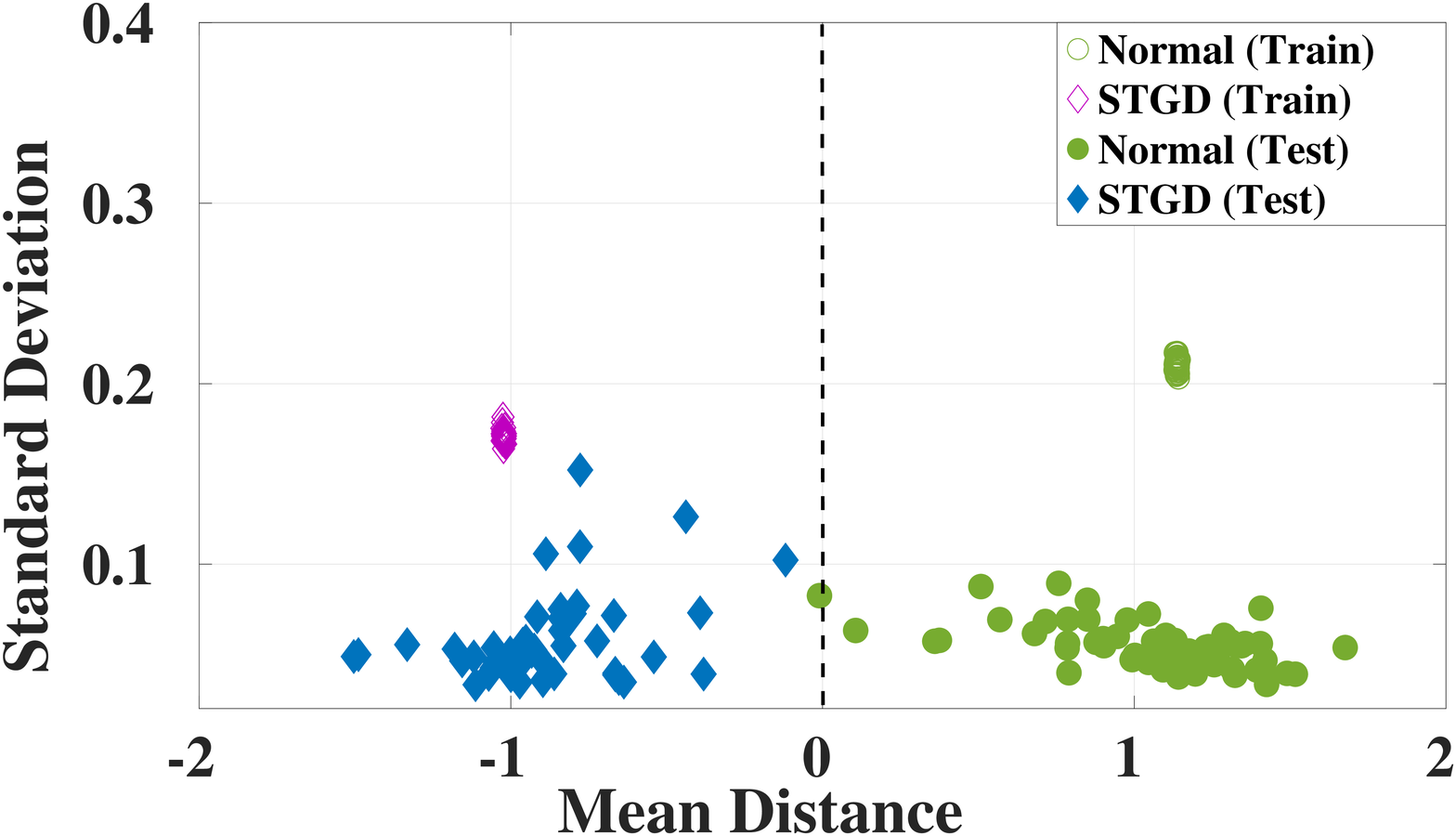} \\
 \\
 (a) & (b)\\

 \includegraphics[width=1\columnwidth]{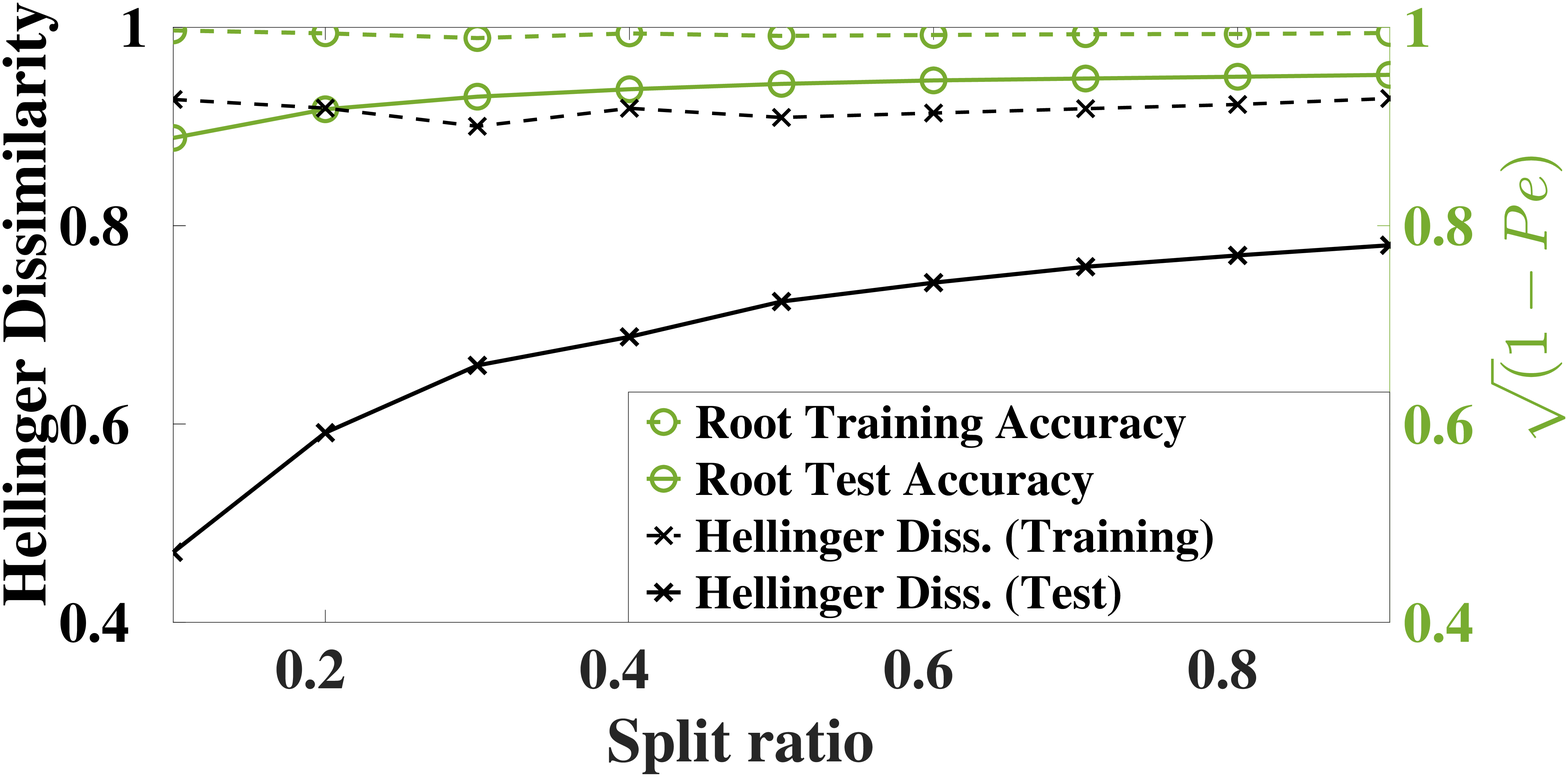} &
 \includegraphics[width=1\columnwidth]{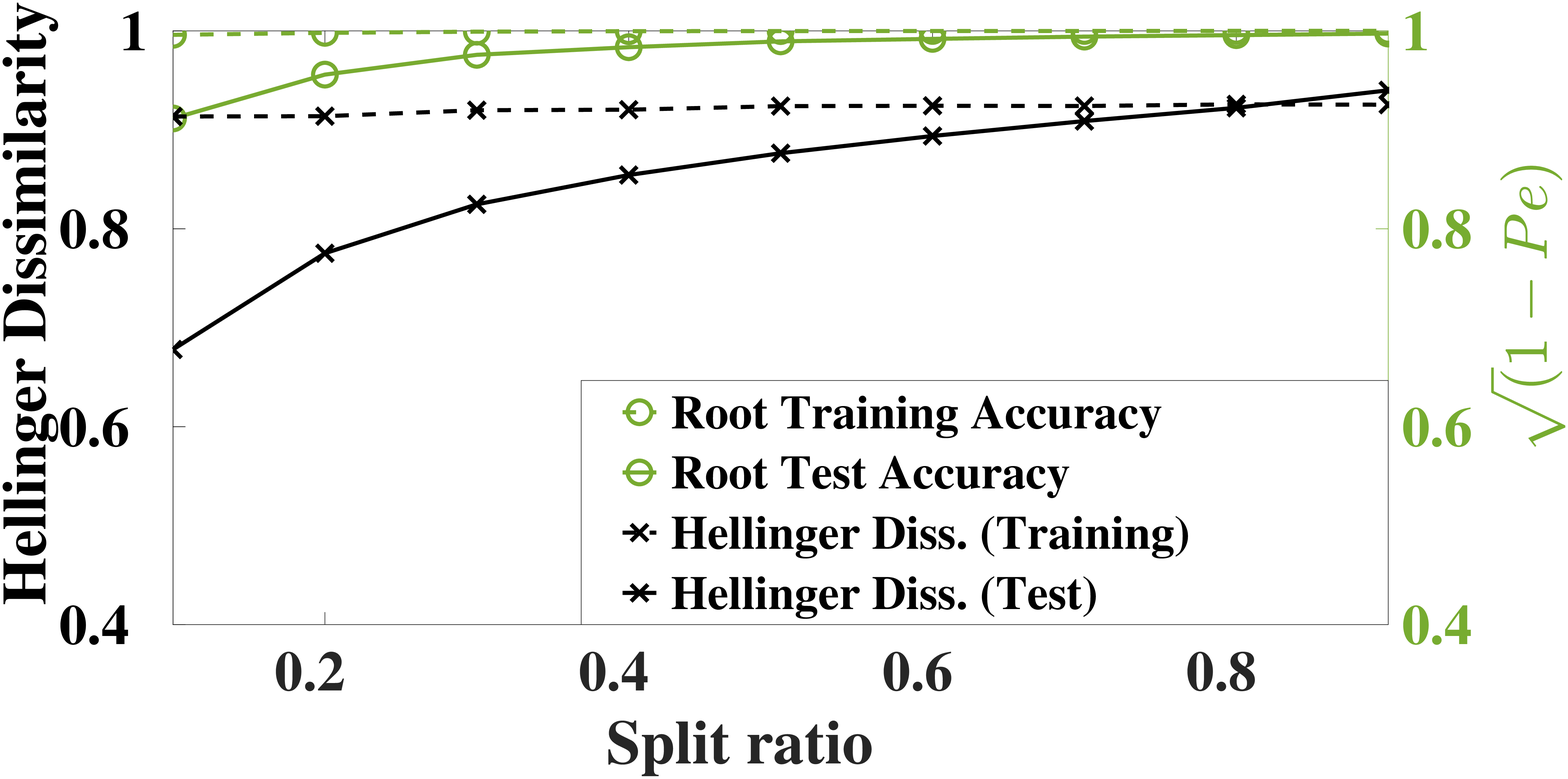}\\
(c) & (d)\\

\end{tabular}
\caption{Signed distances from decision boundary over all iterations of MCCV for RBF SVM on (a) original dataset, and (b) reduced dataset; Variation of Hellinger dissimilarity and root accuracy against split ratio on (c) original dataset, and (d) reduced dataset.}
\label{fig:ObservFig}

\end{figure*}

As alluded earlier, we envisage an SVM-based progress monitoring system, where FAF images of an eye are taken over multiple patient visits. In this setting, one would consider the diagnosed disease, and compute the time course of signed (positive for normal; negative for diseased class) distances of the corresponding feature vectors from the disease-specific decision boundary based on successive FAF images.
For a correctly diagnosed subject, the trajectory should begin within the diseased class. Further, a trajectory heading towards the decision boundary would indicate gradual improvement, while that moving further away would indicate deteriorating condition. Here, while making a feasibility study of the envisaged system for disease progress monitoring in terms of class separation, we restrict only to RBF-SVM classifiers. 

To this end, we first revisit the proposed RBF-SVM screening system, and set the corresponding class separation as reference. In particular, we obtained for each fixed split ratio the optimum decision boundary corresponding to each of the 5000 random partitions mentioned in Section \ref{sec: RES}. Further, fixing each such boundary, we computed the signed (euclidean) distance of each feature vector (corresponding to a FAF image from the training as well as the test subset) from that boundary. Next, corresponding to each FAF image (data point), we obtained a collection of 5000 such distances, and calculated the mean and the standard deviation. In Fig. \ref{fig:ObservFig}(a), we considered the fixed split-ratio 80:20, and obtained a scatter plot of the aforementioned quantities by sweeping the said data point through the entire screening dataset at hand. Here, recall again that our screening dataset consists of FAF images of healthy eyes as well as eyes affected by STGD, CNVM and CSCR. In the above plot, we distinguish between different diseases, only as a visualization aid (without making any disease-specific analysis).

Post diagnosis, the disease affecting each subject would be known (ignoring misdiagnosis), and one potentially needs to monitor subsequent progress of that disease. In this context, the following analysis is presented for the case when the diagnosed disease is STGD in view of its relatively high prevalence. In this regard, we next considered the subset of our dataset that consists of images of only STGD-affected and healthy eyes, and ignored the rest. On this reduced dataset, as earlier, we considered the 80:20 split ratio, and again obtained optimum decision boundaries between diseased (representing only STGD now) and healthy classes using RBF-SVM classifier over 5000 random partitions, and subsequently computed desired distances from respective decision boundaries. Scatter plot of image(data point)-wise mean and standard deviation of those distances are analogously furnished in Fig. \ref{fig:ObservFig}(b). Comparing Fig. \ref{fig:ObservFig}(b) with Fig. \ref{fig:ObservFig}(a), the classes are observed to be more clearly separated, when only STGD-affected eyes represent the diseased class.  More generally, we expect class separation for any disease-specific reduced dataset to be larger than that for the original screening database. 



At this point, we desired to study the above phenomenon more generally across all split ratios. However, as inspection of a series of scatter plots could be subjective and tedious, we now resorted to quantitative comparison of class separation. In particular, we used Hellinger dissimilarity (HD) between probability distributions $p$ and $q$ \cite{deza2006dictionary}, defined by
\begin{equation}
\label{eq:HD}
H = \left({1 - \sum_{x} {p^{\frac{1}{2}}(x)q^{\frac{1}{2}}(x)}}\right)^{\frac{1}{2}}
\end{equation}
as a measure of separation between healthy and diseased classes, by taking $p$ and $q$ to be the respective distributions of distances corresponding to the aforesaid classes, and denoting the distance variable by $x$.

Specifically, for each split ratio, we estimated the probability distribution $p$ (resp. $q$) of aforementioned distances corresponding to healthy (resp. diseased) class over 5000 partitions of training as well as test sets, as customary. Using these distributions, we obtained the HD measure, which is plotted against split ratio for both training and test sets. In particular, presented in Fig. \ref{fig:ObservFig}(c), the aforesaid plots for the original dataset, representing STGD, CNVM and CSCR, show an increasing HD with increasing split ratio for test cases, and a more-or-less constant behavior with slight fluctuations in case of training. Further, HD for training remains greater than that for test throughout. Shown in Fig. \ref{fig:ObservFig}(d), analogous plots for the reduced dataset, representing STGD only, also reveal almost constant HD for training, nearly matching that for the original dataset with even less fluctuations. As expected, the test HD for the reduced dataset, while still exhibiting an increasing behavior, is significantly greater than that for the original dataset for any given split ratio. As an interesting aside, the test HD now crosses over the training HD to become the larger quantity for split ratios slightly greater than 0.8. 


Thus, in view of the significant separation between a specific disease (demonstrated for STGD) and the healthy class, RBF-SVM potentially provides a satisfactory tool for monitoring of disease progress. Of course, here we provide only preliminary evidence, and it remains essential to undertake a confirmatory study involving multiple visits of several post-diagnosis patients in order to draw firm conclusions.

\section{DISCUSSION AND FUTURE COURSE}
\label{sec:future}

In this paper, we developed a screening technique that does not require the participation of ophthalmologists, and semi-automatically identifies the presence of disease based on FAF images with minimal assistance from optometrists. Specifically, we proposed an RBF-SVM classifier, and demonstrated its efficacy in terms of not only classification accuracy, but also other desirable properties. Such a technique holds vast potential in transforming eyecare in remote regions, especially, in developing countries such as India. Indeed, existing organizations, including L V Prasad Eye Institute (LVPEI), which has hierarchical organizational structure with primary centers serving far-flung areas, could be of particular help in propagating the proposed screening service. In turn, such service, apart from screening remote subjects, can potentially make the said organizations more efficient, by referring only medically relevant subjects for diagnosis and/or treatment at secondary or tertiary centers. Further, in view of the excellent class separation exhibited by the proposed classifier, our screening service can potentially be integrated with post-diagnosis treatment response monitoring. In this regard, the distances so far have been measured from the partition-specific decision boundary, the main challenge lies in selecting a representative partition-independent boundary. However, noting low variability in MCCV performance, we believe that the choice of such a representative boundary should not be very difficult. Finally, while FAF imaging presents certain broad advantages over FP, the latter has benefits of its own. Accordingly, in future, we plan to incorporate FP alongside FAF imaging to improve screening (as well as disease progress monitoring) performance.

%
%

\section*{Acknowledgements}
The work was partly supported by Grant BT/PR16582/BID/7/667/2016, Department of Biotechnology (DBT), Ministry of Science and Technology, the Government of India. Shanmukh Reddy Manne thanks the Ministry of Electronics and Information
Technology (MeitY), the Government of India, for fellowship grant under Visvesvaraya PhD Scheme for Electronics and IT.

\appendix

\section{Theoretical Connection}
\label{sec:app}

While development and demonstration of a screening tool remained our primary focus, we seek to point out an information-theoretic connection that may interest certain readers. To begin with, revisit 
Figs. \ref{fig:ObservFig}(c) and \ref{fig:ObservFig}(d), and consider the variation in the square root of classification accuracy levels in various cases, distinguishing between the training and test scenarios.
Comparing analogous cases, we observe that both the root classification accuracy and the HD measure exhibit somewhat similar variational behavior, although the former always remains greater than the latter. 

This observation is in fact consistent with an information-theoretic principle. To see this, consider the well-known Chernoff bound
\begin{equation}
\label{eq:CB}
P_e < \sum_{x} p^{1-s}(x)q^{s}(x),
\end{equation}
which holds for any exponent $0<s<1$ \cite{Poor}. Taking $s=\frac{1}{2}$, using definition (\ref{eq:HD}) of HD, and reorganizing (\ref{eq:CB}), we obtain 
\begin{equation}
\label{eq:Bound}
H<\sqrt{1-P_e}.
\end{equation}
The bound (\ref{eq:Bound}) explains the observed dominance of root classification accuracy over HD. 

\begin{figure}[t!]
\centering
 \includegraphics[scale = 0.15]{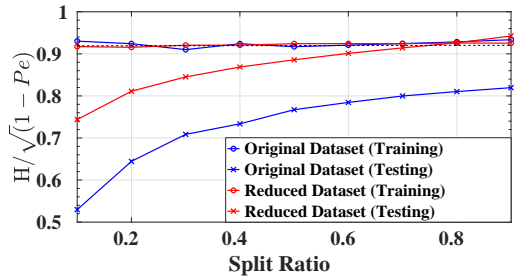}
\caption{Variation of fraction $H/\sqrt{1-P_e}$ against split ratio on both the datasets.}
\label{fig:Frac}

\end{figure}

Further, the proposed RBF-SVM classifier attempts to maximize the class separation in the training subset for each split ratio, bringing $H$ and $\sqrt{1-P_e}$ close. In fact, for training, we observe the fraction $H/\sqrt{1-P_e}$ to hover around a constant 0.92 across various split ratios, for both the original and the reduced datasets (see Fig. \ref{fig:Frac}). This indicates that the complexity of maximization of class separation which underlies training remains similar for competing datasets across split ratios. However, in the test scenarios, such fraction exhibits an increasing behavior against split ratio, indicating that the gap between $H$ and $\sqrt{1-P_e}$ in (\ref{eq:Bound}) narrows when higher proportion of training data becomes available, as anticipated. Further, the said fraction for the original dataset is found to be smaller than that for the reduced dataset for every split ratio, reflecting the decrease in problem complexity of the latter disease-specific problem. In fact, for the reduced dataset, the test fraction crosses the training fraction for large split ratios.





\bibliography{refs}

\begin{thebibliography}{10}
\expandafter\ifx\csname url\endcsname\relax
  \def\url#1{\texttt{#1}}\fi
\expandafter\ifx\csname urlprefix\endcsname\relax\def\urlprefix{URL }\fi
\expandafter\ifx\csname href\endcsname\relax
  \def\href#1#2{#2} \def\path#1{#1}\fi

\bibitem{yolcu2014imaging}
U.~Yolcu, O.~F. Sahin, F.~C. Gundogan, Imaging in ophthalmology, in:
  Ophthalmology-Current Clinical and Research Updates, IntechOpen, 2014.

\bibitem{nickla2010multifunctional}
D.~L. Nickla, J.~Wallman, The multifunctional choroid, Progress in retinal and
  eye research 29~(2) (2010) 144--168.

\bibitem{schmitz2008fundus}
S.~Schmitz-Valckenberg, F.~G. Holz, A.~C. Bird, R.~F. Spaide, Fundus
  autofluorescence imaging: review and perspectives, Retina 28~(3) (2008)
  385--409.

\bibitem{AMD_STGD}
R.~T. Smith, N.~Lee, J.~Chen, M.~Busuioc, A.~F. Laine, Interactive image
  analysis in age-related macular degeneration (amd) and stargardt disease
  (stgd), in: 2008 42nd Asilomar Conference on Signals, Systems and Computers,
  IEEE, 2008, pp. 651--654.

\bibitem{Sect_Stargardt}
T.~R. Burke, T.~Duncker, R.~L. Woods, J.~P. Greenberg, J.~Zernant, S.~H. Tsang,
  R.~T. Smith, R.~Allikmets, J.~R. Sparrow, F.~C. Delori, Quantitative fundus
  autofluorescence in recessive stargardt disease, Investigative ophthalmology
  \& visual science 55~(5) (2014) 2841--2852.

\bibitem{Retina2013}
M.~Marsiglia, S.~Boddu, S.~Bearelly, L.~Xu, B.~E. Breaux, K.~B. Freund, L.~A.
  Yannuzzi, R.~T. Smith, Association between geographic atrophy progression and
  reticular pseudodrusen in eyes with dry age-related macular degeneration,
  Investigative ophthalmology \& visual science 54~(12) (2013) 7362--7369.

\bibitem{schmitz2008evaluation}
S.~Schmitz-Valckenberg, M.~Fleckenstein, A.~P. G{\"o}bel, K.~Sehmi, F.~W.
  Fitzke, F.~G. Holz, A.~Tufail, Evaluation of autofluorescence imaging with
  the scanning laser ophthalmoscope and the fundus camera in age-related
  geographic atrophy, American journal of ophthalmology 146~(2) (2008)
  183--192.

\bibitem{zola2018evolution}
M.~Zola, I.~Chatziralli, D.~Menon, R.~Schwartz, P.~Hykin, S.~Sivaprasad,
  Evolution of fundus autofluorescence patterns over time in patients with
  chronic central serous chorioretinopathy, Acta ophthalmologica 96~(7) (2018)
  e835--e839.

\bibitem{jolly2016novel}
J.~K. Jolly, S.~K. Wagner, J.~Moules, F.~Gekeler, A.~R. Webster, S.~M. Downes,
  R.~E. MacLaren, A novel method for quantitative serial autofluorescence
  analysis in retinitis pigmentosa using image characteristics, Translational
  vision science \& technology 5~(6) (2016) 10--10.

\bibitem{schmitz2009fundus}
S.~Schmitz-Valckenberg, M.~Fleckenstein, H.~P. Scholl, F.~G. Holz, Fundus
  autofluorescence and progression of age-related macular degeneration, Survey
  of ophthalmology 54~(1) (2009) 96--117.

\bibitem{schmitz2011semiautomated}
S.~Schmitz-Valckenberg, C.~K. Brinkmann, F.~Alten, P.~Herrmann, N.~K.
  Stratmann, A.~P. G{\"o}bel, M.~Fleckenstein, M.~Diller, G.~J. Jaffe, F.~G.
  Holz, Semiautomated image processing method for identification and
  quantification of geographic atrophy in age-related macular degeneration,
  Investigative ophthalmology \& visual science 52~(10) (2011) 7640--7646.

\bibitem{fleckenstein2011fundus}
M.~Fleckenstein, S.~Schmitz-Valckenberg, C.~Martens, S.~Kosanetzky, C.~K.
  Brinkmann, G.~S. Hageman, F.~G. Holz, Fundus autofluorescence and
  spectral-domain optical coherence tomography characteristics in a rapidly
  progressing form of geographic atrophy, Investigative ophthalmology \& visual
  science 52~(6) (2011) 3761--3766.

\bibitem{holz2007progression}
F.~G. Holz, A.~Bindewald-Wittich, M.~Fleckenstein, J.~Dreyhaupt, H.~P. Scholl,
  S.~Schmitz-Valckenberg, F.-S. Group, et~al., Progression of geographic
  atrophy and impact of fundus autofluorescence patterns in age-related macular
  degeneration, American journal of ophthalmology 143~(3) (2007) 463--472.

\bibitem{fleckenstein2018progression}
M.~Fleckenstein, P.~Mitchell, K.~B. Freund, S.~Sadda, F.~G. Holz, C.~Brittain,
  E.~C. Henry, D.~Ferrara, The progression of geographic atrophy secondary to
  age-related macular degeneration, Ophthalmology 125~(3) (2018) 369--390.

\bibitem{schuerch2017quantifying}
K.~Schuerch, R.~L. Woods, W.~Lee, T.~Duncker, F.~C. Delori, R.~Allikmets, S.~H.
  Tsang, J.~R. Sparrow, Quantifying fundus autofluorescence in patients with
  retinitis pigmentosa, Investigative ophthalmology \& visual science 58~(3)
  (2017) 1843--1855.

\bibitem{acharya2016automated}
U.~R. Acharya, M.~R.~K. Mookiah, J.~E. Koh, J.~H. Tan, S.~V. Bhandary, A.~K.
  Rao, H.~Fujita, Y.~Hagiwara, C.~K. Chua, A.~Laude, Automated screening system
  for retinal health using bi-dimensional empirical mode decomposition and
  integrated index, Computers in biology and medicine 75 (2016) 54--62.

\bibitem{iovs2007comparing}
S.~Bearelly, M.~Chen, M.~Kelly, S.~Cousins, Comparing fundus autofluorescence
  imaging with fundus photographs in patients with geographic atrophy related
  to age-related macular degeneration, Investigative Ophthalmology \& Visual
  Science 48~(13) (2007) 2163--2163.

\bibitem{retina2012comparison}
A.~A. Khanifar, D.~E. Lederer, J.~H. Ghodasra, S.~S. Stinnett, J.~J. Lee, S.~W.
  Cousins, S.~Bearelly, Comparison of color fundus photographs and fundus
  autofluorescence images in measuring geographic atrophy area, Retina
  (Philadelphia, Pa.) 32~(9) (2012) 1884.

\bibitem{hadi2013comparison}
A.~M.~A. Hadi, K.~G. Andrawos, et~al., Comparison between fundus
  autofluorescence images and color fundus photos in patients with late dry
  age-related macular degeneration, Egyptian Retina Journal 1~(3) (2013) 31.

\bibitem{FAF_clinical}
M.~Yung, M.~A. Klufas, D.~Sarraf, Clinical applications of fundus
  autofluorescence in retinal disease, International journal of retina and
  vitreous 2~(1) (2016) 12.

\bibitem{FAF_CNV}
S.~S. Dandekar, S.~A. Jenkins, T.~Peto, H.~P. Scholl, K.~S. Sehmi, F.~W.
  Fitzke, A.~C. Bird, A.~R. Webster, Autofluorescence imaging of choroidal
  neovascularization due to age-related macular degeneration, Archives of
  Ophthalmology 123~(11) (2005) 1507--1513.

\bibitem{early1991early}
E.~T. D. R. S.~R. Group, et~al., Early photocoagulation for diabetic
  retinopathy: Etdrs report number 9, Ophthalmology 98~(5) (1991) 766--785.

\bibitem{cortes1995support}
C.~Cortes, V.~Vapnik, Support-vector networks, Machine learning 20~(3) (1995)
  273--297.

\bibitem{LIBSVM}
C.-C. Chang, C.-J. Lin, Libsvm: A library for support vector machines, ACM
  transactions on intelligent systems and technology (TIST) 2~(3) (2011) 27.

\bibitem{xu2001monte}
Q.-S. Xu, Y.-Z. Liang, Monte carlo cross validation, Chemometrics and
  Intelligent Laboratory Systems 56~(1) (2001) 1--11.

\bibitem{milk}
S.~Tripathy, M.~S. Reddy, S.~R.~K. Vanjari, S.~Jana, S.~G. Singh, A step
  towards miniaturized milk adulteration detection system: Smartphone-based
  accurate ph sensing using electrospun halochromic nanofibers, Food Analytical
  Methods 12~(2) (2019) 612--624.

\bibitem{macek2008}
K.~Macek, Pareto principle in datamining: an above-average fencing algorithm,
  Acta Polytechnica 48~(6) (2008).

\bibitem{LDA}
Z.~Qiao, L.~Zhou, J.~Z. Huang, Effective linear discriminant analysis for high
  dimensional, low sample size data, in: Proceeding of the world congress on
  engineering, Vol.~2, Citeseer, 2008, pp. 2--4.

\bibitem{DQAembc19}
C.~Dev, M.~Sharang, S.~R. Manne, A.~Goud, S.~B. Bashar, A.~Richhariya,
  J.~Chhablani, K.~K. Vupparaboina, S.~Jana, Diagnostic quality assessment of
  ocular fundus photographs: Efficacy of structure-preserving scatnet features,
  in: 2019 41st Annual International Conference of the IEEE Engineering in
  Medicine and Biology Society (EMBC), IEEE, 2019, pp. 2091--2094.

\bibitem{xgboost}
T.~Chen, C.~Guestrin, Xgboost: A scalable tree boosting system, in: Proceedings
  of the 22nd acm sigkdd international conference on knowledge discovery and
  data mining, ACM, 2016, pp. 785--794.

\bibitem{SRFPEDembc19}
R.~V. Teja, S.~R. Manne, A.~Goud, M.~A. Rasheed, K.~K. Dansingani,
  J.~Chhablani, K.~K. Vupparaboina, S.~Jana, Classification and quantification
  of retinal cysts in oct b-scans: Efficacy of machine learning methods, in:
  2019 41st Annual International Conference of the IEEE Engineering in Medicine
  and Biology Society (EMBC), IEEE, 2019, pp. 48--51.

\bibitem{RF}
L.~Breiman, Random forests, Machine learning 45~(1) (2001) 5--32.

\bibitem{MLtools}
P.~Samant, R.~Agarwal, Machine learning techniques for medical diagnosis of
  diabetes using iris images, Computer methods and programs in biomedicine 157
  (2018) 121--128.

\bibitem{ICIP2015}
S.~Morales, K.~Engan, V.~Naranjo, A.~Colomer, Detection of diabetic retinopathy
  and age-related macular degeneration from fundus images through local binary
  patterns and random forests, in: 2015 IEEE International Conference on Image
  Processing (ICIP), IEEE, 2015, pp. 4838--4842.

\bibitem{deza2006dictionary}
M.-M. Deza, E.~Deza, Dictionary of distances, Elsevier, 2006.

\bibitem{Poor}
H.~V. Poor, An introduction to signal detection and estimation, Springer
  Science \& Business Media, 2013.

\end{thebibliography}

\end{document}